\pdfoutput=1
\documentclass[11pt]{article}

\usepackage[]{ACL2023}

\usepackage{times}
\usepackage{latexsym}
\usepackage{amsmath}
\usepackage{amssymb}
\usepackage{subcaption}

\DeclareMathOperator*{\argmax}{argmax}
\usepackage[T1]{fontenc}
\usepackage[utf8]{inputenc}

\usepackage{microtype}

\usepackage{inconsolata}

\usepackage{multirow}
\usepackage{floatrow}
\usepackage{graphicx}
\usepackage{array}
\usepackage{caption}
\usepackage{subcaption}

\definecolor{amber(sae/ece)}{rgb}{1.0, 0.49, 0.0}
\definecolor{burntorange}{rgb}{0.8, 0.33, 0.0}
\definecolor{dukeblue}{rgb}{0.0, 0.0, 0.61}
\title{Vision Meets Definitions: Unsupervised Visual Word Sense Disambiguation Incorporating Gloss Information}

\author{\begin{tabular}[c]{@{}c@{}}Sunjae Kwon$^{1}$, Rishabh Garodia$^{1}$, Minhwa Lee$^{1}$, Zhichao Yang$^{1}$, Hong Yu$^{1,2,3,4}$\end{tabular} \\
  \begin{tabular}[c]{@{}c@{}} 
  $^{1}$UMass Amherst, $^{2}$UMass Lowell, $^{3}$UMass Chan Medical School, $^{4}$ VA Bedford Health Care
  \end{tabular}
  \\
  \begin{tabular}[c]{@{}c@{}}
  \texttt{sunjaekwon@umass.edu, rgarodia@umass.edu, minhwalee@umass.edu}\\\texttt{zhichaoyang@umass.edu, hong\_yu@uml.edu}\end{tabular} \\}

\begin{document}
\maketitle
\begin{abstract}
Visual Word Sense Disambiguation (VWSD) is a task to find the image that most accurately depicts the correct sense of the target word for the given context. Previously, image-text matching models often suffered from recognizing polysemous words. This paper introduces an unsupervised VWSD approach that uses gloss information of an external lexical knowledge-base, especially the sense definitions. Specifically, we suggest employing Bayesian inference to incorporate the sense definitions when sense information of the answer is not provided. In addition, to ameliorate the out-of-vocabulary (OOV) issue, we propose a context-aware definition generation with GPT-3. Experimental results show that VWSD performance increased significantly with our Bayesian inference-based approach. In addition, our context-aware definition generation achieved prominent performance improvement in OOV examples exhibiting better performance than the existing definition generation method.
%our suggestions 

\end{abstract}

\section{Introduction}
With the development of deep learning technology, research on multimodality such as Visio-Linguistic Models (VLMs) has been actively conducted \citep{schneider2022golden}. 
In particular, state-of-the-art VLMs, such as image-text matching (ITM) models \citep{radford2021learning, singh2022flava} and text-to-image generation models \citep{rombach2022high,seneviratne2022dalle}, are employed in many industrial projects, including image retrieval systems \citep{yuan2021conversational,yuan2021exploring} and AI-assisted image generators \citep{das2022explaining, seneviratne2022dalle}.

% However, several recent studies claim that VLMs frequently fail on recognizing and mapping the correct sense of ambiguous words \citep{rassin2022dalle}. This is because, usually text-image models do not consider the sense of each ambiguous word while they are pretrained. Thus, it is important to 

\begin{figure}[!ht]
    \centering
    \includegraphics[width=.7\textwidth]{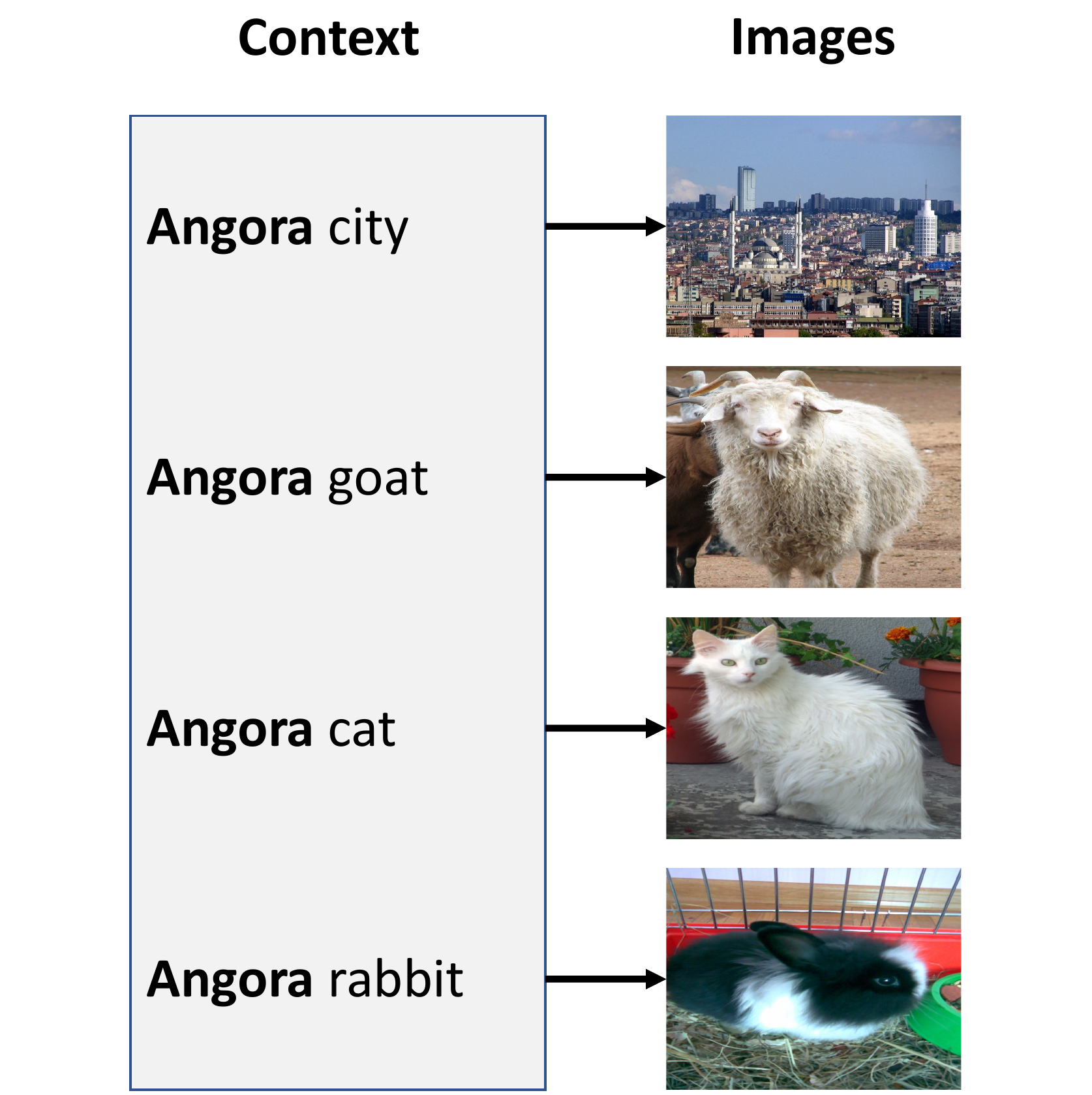}
    \caption{An example of VWSD from SemEval-2023 task 1 dataset \citep{raganato-etal-2023-semeval}. We can see that even if the target word (`\textbf{Angora}') is the same, different images should be selected according to the context.}
    \label{fig:VWSD_example}
\end{figure}
% (ACL 결과 언제 나오죠? ㅋㅋㅋㅋ) 

Visual Word Sense Disambiguation (VWSD) is a multimodal task of natural language processing (NLP) and computer vision that selects the image which corresponds to the intended meaning of the target word among a set of candidate images \citep{raganato-etal-2023-semeval}.  Figure~\ref{fig:VWSD_example} is an example of VWSD. For the ambiguous target word \footnote{An ambiguous word that we want to disambiguate with machines.} `Angora', we can notice that the answer image should be conditionally changed regarding the context. VWSD can play an important role in several downstream tasks including image retrieval \citep{chen2015sense}, action recognition \citep{gella2017disambiguating} and visual question answering \citep{whitehead2020learning}.

\begin{figure}
    \centering
    \includegraphics[width=\linewidth]{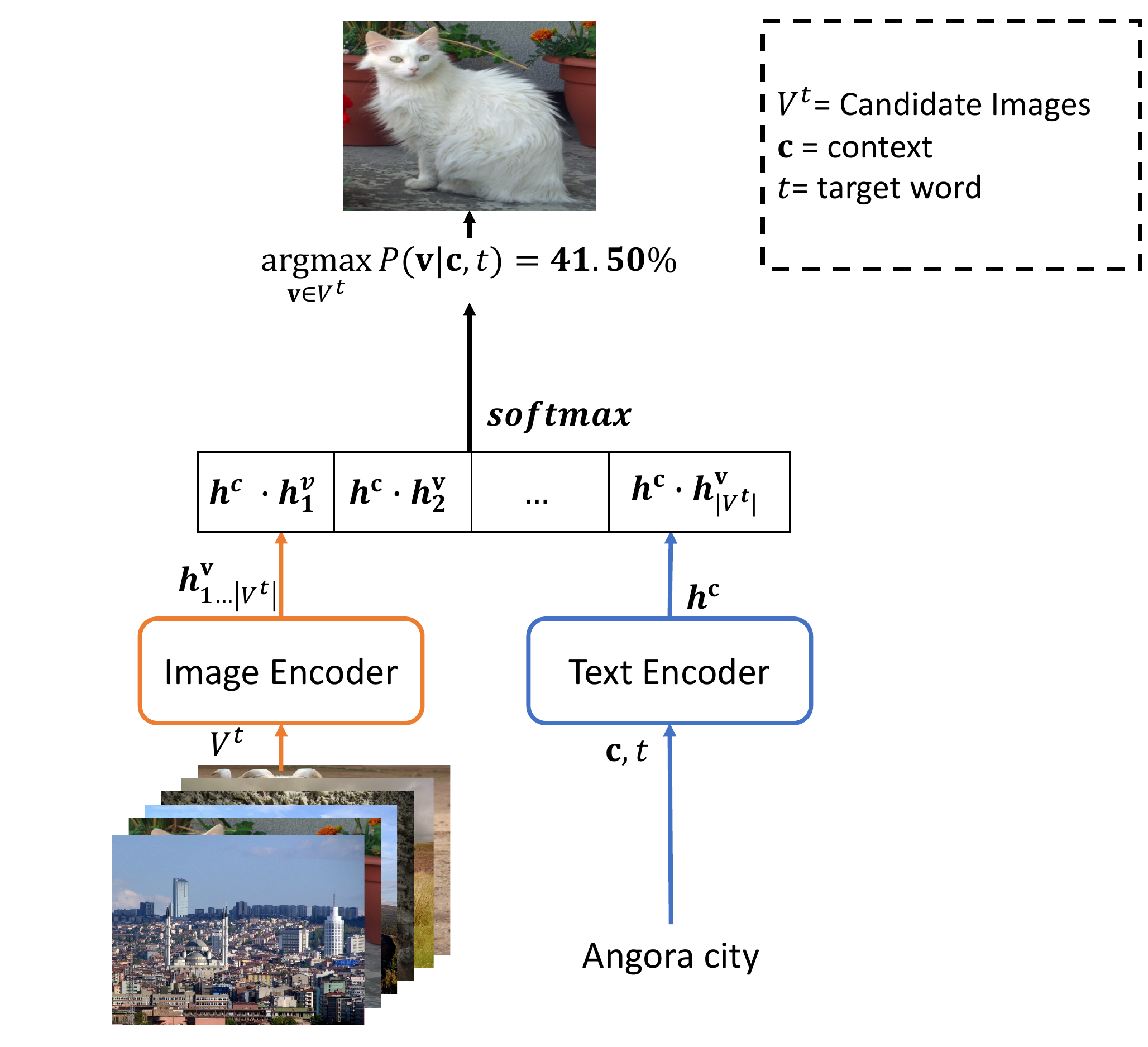}
    \caption{Illustrative concepts and an example input on a CLIP model \citep{radford2021learning}.}
    \label{fig:clip_structure}
\end{figure}

Unsupervised VWSD can be formulated in the same way as the ITM task \citep{cao2022image}, that is, finding the images that best match the given context. However, VWSD often requires more complex reasoning on both text and images than conventional ITM models. The example in Figure~\ref{fig:clip_structure} demonstrates that CLIP \citep{radford2021learning}, a state-of-the-art (SOTA) ITM model, fails to recognize the answer image for the given context \footnote{Text surrounding a target word which is used as a clue to disambiguate the target word (e.g. Angola cat, Angola city, Angola goat in Figure 1).}. This limitation of VLMs, where they fail to handle ambiguous words, was also reported in another study on an image generation model \citep{rassin2022dalle}.

%To ameliorate this issue, this paper introduces a novel unsupervised 
%VWSD 
%approach that ~ inference using the definition descriptions in the glossary of lexical knowledge-bases without any further training. Especially, ~. 
To ameliorate this problem, we propose to disambiguate visual words with the assistance of a glossary of lexical knowledge-bases (LKBs) without the use of any further training or dataset. Specifically, we utilize the sense definitions of an ambiguous word that have been widely exploited in previous lexical semantic tasks \citep{raganato2017word, gella2017disambiguating, pilehvar2019wic}. Herein, since the answer sense of the target word is not provided in the VWSD setting, we propose an approach derived from Bayesian inference, using pretained ITM models. Moreover, in order to deal with out-of-vocabulary (OOV) words that cannot find the sense definitions of the target word in LKBs, we suggest the concept of context-aware definition generation (CADG). The definitions of a target word are generated by a large language model, GPT-3 \citep{brown2020language}, as auxiliary information for VWSD. % In particular, we design the context-aware prompt for the definition generation which did not consider in the previous work \citep{malkin2021gpt}. 
%Herein, by following \citet{malkin2021gpt}'s approach, prompts input to GPT-3 \citep{brown2020language} to generate the definition of the given target word.

Experiments were conducted on SemEval-2023 (SE23) Task 1-Visual-WSD \citep{raganato-etal-2023-semeval}, a publicly available VWSD dataset. Furthermore, in the experiments, we utilized two pretained SOTA ITM models: (1) CLIP \citep{radford2021learning} and (2) FLAVA \citep{singh2022flava}. 
Experiments showed that our proposed approach significantly improved the performance of baseline ITM models. In addition, we demonstrated that our concept of CADG not only significantly increased the performance of OOV cases but is also more advantageous than the previous definition generation approach. We implement experimental codes in 
\url{https://github.com/soon91jae/UVWSD}.

The contributions of this paper can be summarized as follows: 

\begin{itemize}
    \item This paper introduces a new gloss-incorporated VWSD approach inspired by Bayesian inference. 
    \item Experimental results show that our Bayesian inference-based approach boosted the unsupervised VWSD performance significantly without any additional training. 
     \item Furthermore, we suggest the CADG method to challenge the OOV issue.   
    
    % \item 
    %e method further improved the performance. 
    %compared to the previous definition generation approach.
    %\item Providing qualitative analyses on the 
\end{itemize}
\begin{figure*}
    \centering
    \includegraphics[height=0.32\textheight]{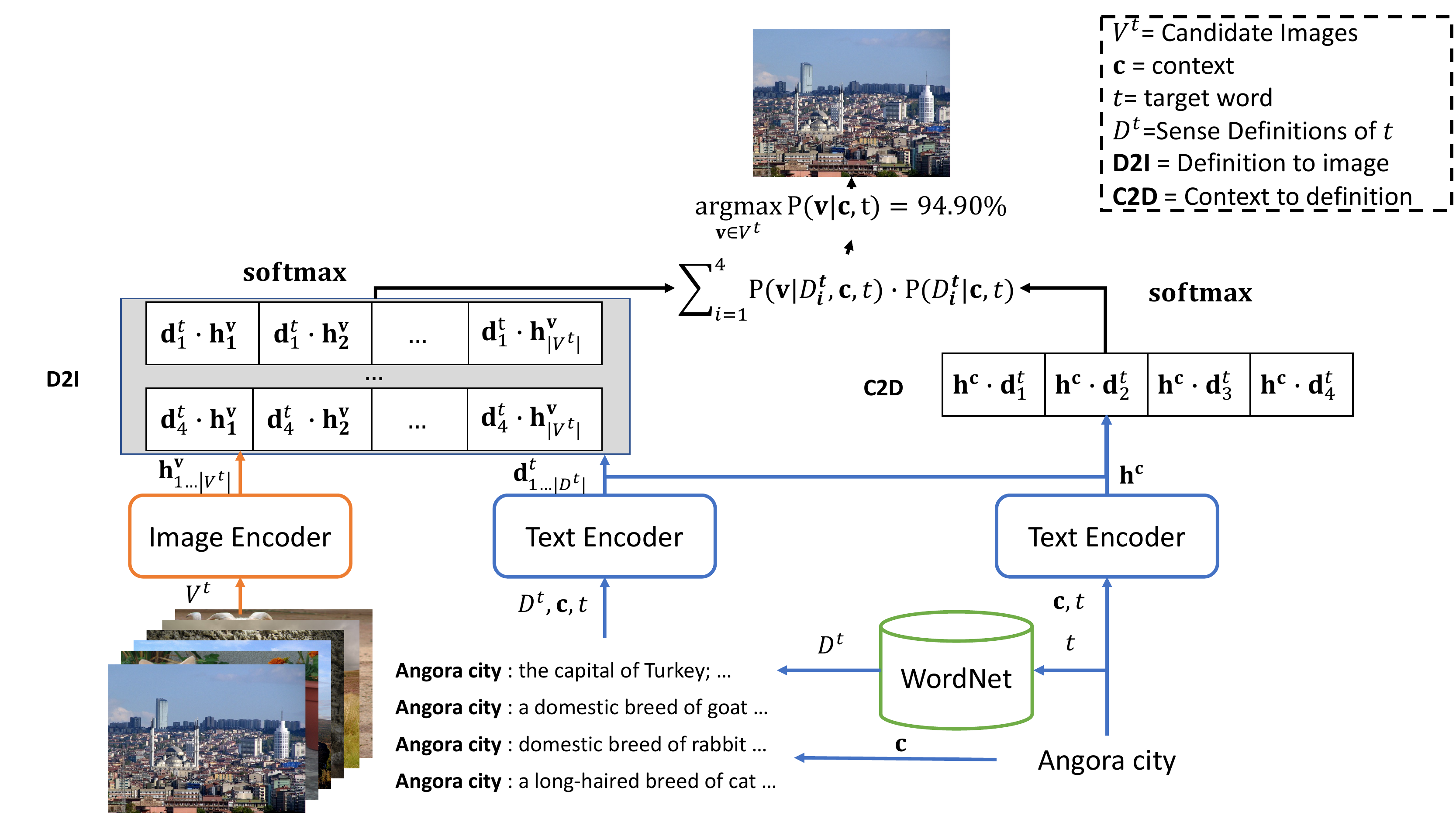} %\includegraphics[width=\textwidth]
    \caption{Illustrative concepts and an example input on our gloss-enhanced framework on a CLIP model. Note that, even though the image encoder and the text encoder are the exactly same as those in Figure~\ref{fig:clip_structure}, our approach can correctly predict the answer image different from the original CLIP model's prediction.}
    \label{fig:our_model_structure}
\end{figure*}
\section{Related Work}
\subsection{Word and Visual Sense Disambiguation}
VWSD task is closely relevant to a line of sense disambiguation studies. One of them is Word Sense Disambiguation (WSD) which automatically identifies ambiguous words into corresponding senses \citep{dongsuk2018word}. The early stage of WSD research tried to employ diverse information in LKBs with unsupervised manners such as lexical similarity \citep{kilgarriff2000english}, knowledge-graph connectivity \citep{agirre2014random, kwon2021word}, and topic modeling \citep{chaplot2018knowledge}. After the emergence of pretrained language models (LMs) such as BERT \citep{devlin-etal-2019-bert}, LM-based transfer learning approaches have been actively studied \citep{huang2019glossbert, barba2021consec}. In particular, gloss-enhanced WSD models that use sense definition and context together using a cross-encoder \citep{huang2019glossbert, barba2021esc} or bi-encoder \citep{blevins2020moving} structures are not only overwhelm existing approaches but also robust against few-shot examples. \citet{wahle2021incorporating} suggest incorporating WordNet knowledge into LMs while pre-training them. Specifically, the authors utilize a multi-task learning method that trains LMs with both mask language modeling loss and WSD task loss.   

Visual Verb Sense Disambiguation (VVSD) is another task relevant to VWSD. VVSD is a multimodal sense disambiguation task that selects the correct sense of a pair of a given ambiguous verb word and image \citep{gella2017disambiguating}. \citet{gella2017disambiguating} suggest an unsupervised VVSD approach that takes advantage of various Visio-linguistic features (image representation, object label, image caption features) together and calculates the matching score between an image and a sense definition with a variant of Lesk algorithm. \citet{vascon2021transductive} propose a semi-supervised VVSD method based on game theoretic transduction for inference. %Transductive Visual Verb Sense Disambiguation
Meanwhile, \citet{gella2019cross} demonstrate that a VVSD model trained on multi-lingual VVSD dataset not only benefit the performance on verb sense disambiguation but also boost the performance of a downstream task, the multi-modal machine translation task.  

Our work is related to gloss-enhanced WSD models in that we are using both sense definition and context together. However, our study differs from previous WSD studies in that it tackles a multi-modal task. It is also relevant to VVSD in terms of multi-modal sense disambiguation. However, VVSD systems \citep{gella2016unsupervised} are usually designed to analyze a small number of verb words, while the VWSD task contains a lot of nouns and adjectives. Finally, our work tackles a new VWSD task and we introduce a method of implementing sense definitions with SOTA ITM models based on Bayesian inference where sense definitions as a latent variable.

\subsection{Definition Generation} 
Our CADG is related to the definition generation task introduced by \citet{noraset2017definition}. The purpose of the task is to generate a definition for a given word. \citet{noraset2017definition} suggest utilizing recurrent neural network-based LMs (RNNLMs) with the definitions collected from WordNet and GNU Collaborative International Dictionary of English (GCIDE). \citet{gadetsky2018conditional} propose definition generation models to handle polysemous words with context and the soft-attention mechanism. \citet{li2020explicit} propose to perform semantic decomposition of the meanings of words and then use discrete latent variables to model them to generate definitions. \citet{malkin2021gpt} show that a large language model (GPT-3) could generate definitions of neologisms without additional fine-tuning. Herein, the authors suggest generating neologisms with long short-term memory (LSTM) \citep{yu2019review} and definitions of neologisms with a large pretrained LM, GPT-3 \citep{brown2020language}. 
CADG is similar to the one used by \citet{malkin2021gpt}, which involves generating definitions using GPT-3. However, CADG differs in that it takes context into account when generating prompts. Additionally, this study differs from previous work in that it takes context into account when generating prompts and demonstrates that the definitions produced by CADG can be effectively used in downstream tasks, rather than focusing solely on the definition generation task itself.
% CADG is related to the \citet{malkin2021gpt}'s approach in that it generates definitions based on GPT-3. However, the first difference is that it uses context-considered prompts. We also show that the generated definitions can be usefully adopted in a downstream task.

% \begin{figure*}
%     \centering
%     \begin{subfigure}[b]{.27\textwidth}
%         \centering
        
%          \sidesubfloat[]{\includegraphics[height=0.23\textheight]{Figures/CLIP.pdf}}
%          \label{fig:clip}
%      \end{subfigure}\hspace*{\fill}
%      \begin{subfigure}[b]{.57\textwidth}
%         \centering
%          \sidesubfloat[]{\includegraphics[height=0.23\textheight]{Figures/model structure.pdf}}
%          \label{fig:ours}
%      \end{subfigure}
%     \caption{Illustrative concepts and an example input to original CLIP (a) and our gloss enhanced on CLIP (b).}
%     \label{fig:model_structures}
% \end{figure*}

\section{Task Definition on Unsupervised VWSD}
We formulate unsupervised VWSD as a multiclass classification task \citep{aly2005survey} as shown in Eq.~\ref{eq:VWSD}. Unlike the image retrieval task \citep{jing2005unified} that ranks the most relevant images for the given text or keyword, VWSD is designed to choose a specific target $t$ in the given context $c$. Specifically, we define the task to find the image $\hat{\mathbf{v}}$ with the highest posterior probability from a set of images $V^{t}$ that consists of one answer image and other distractors on the target word.
\begin{equation}
    \label{eq:VWSD}
    \hat{\mathbf{v}}=\underset{\mathbf{v} \in V^{t}}{\argmax}~P(\mathbf{v}|\mathbf{c},t)
\end{equation}

Any pretrained ITM models (e.g., CLIP) can calculate the posterior. In Figure~\ref{fig:clip_structure}, a set of candidate images $V^t$ is entered into the image encoder for the target word $t$. At the same time, the context $\mathbf{c}$ that includes $t$ as a part is entered into the text encoder. Then, the inner product of the output hidden representations on images $\mathbf{h}^v_{1...|V^t|}$ and the context $\mathbf{h}^c$ are input to softmax function, which then computes a probability distribution over the images. Finally, the image that produces the highest probability will be selected as the prediction of the model for the target $t$, provided the context $\mathbf{c}$. 

\section{Unsupervised VWSD Incorporating Gloss Information}
Usually, zero-shot ITM models are pretrained without much consideration of polysemous words. For example, Figure~\ref{fig:clip_structure} demonstrates that CLIP fails to predict the correct answer for the target word `Angora', although it is provided with a clear hint of `city' in the given context. Therefore, the zero-shot performance of pretrained ITM models may be limited in the VWSD task. One solution is to use gloss information of a lexical knowledge-base (LKB), particularly exploiting sense definitions. This is because the definitions in LKBs elaborate on each sense for readers who do not know the meaning. Thus, we assume that the sense definitions in LKBs can boost ITM models to conduct VWSD, by injecting the meaning of the correct sense on the input of these models. However, since there is no correct sense information for the target word, it is difficult to apply it directly. For this reason, we suggest a novel gloss-incorporated VWSD approach inspired by Bayesian inference, as presented in Eq.~\ref{eq:bayesian_VWSD}.

Suppose $D^t$ is a set of definitions for the target word $t$ extracted from an LKB. Herein, by using the chain rule, the posterior can be divided into two conditional probabilities associated with a latent variable $D^t$.
\begin{equation}
    P(\mathbf{v}|\mathbf{c},t)=\sum_{i=1}^{|D^t|} P(\mathbf{v}|D^t_i,\mathbf{c},t)P(D^t_i|\mathbf{c},t)
    \label{eq:bayesian_VWSD}
%\frac{P(I,\mathbf{t})}{P(\mathbf{t})}
\end{equation}
In this case, the right term $P(D^t_i|\mathbf{c},t)$ (Context to Definition; C2D) is predicting the conditional probability over the given $i$th sense definition $D^t_i$ for the given target word $t$ and context $\textbf{c}$ which is similar to the gloss-enhanced WSD models \citep{huang2019glossbert, blevins2020moving}. Meanwhile, the left term $P(\mathbf{v}|D^t_i,\mathbf{c},t)$ (Definition to Image; D2I) is the conditional probability of $\mathbf{v}$ for a given the $i$th sense definition, the context and the target word. In doing so, it allows for the development of sophisticated ITMs by enriching the context with its relevant sense definition. Finally, we can calculate $P(\mathbf{v}|\mathbf{c},t)$ by marginalizing over all available sense definitions $D^t_{1...|D^t|}$. 

Figure~\ref{fig:our_model_structure} demonstrates an illustrative concept of our gloss-incorporated VWSD approach with a pre-trained CLIP. First, similar to the original CLIP, a set of candidate images $V^t$ and a context $\mathbf{c}$ are input to the image encoder and the text encoder, respectively. Meanwhile, a set of definitions of the target word $D^t$ is extracted from an LKB. In our work, we utilize WordNet \citep{miller1995wordnet} which has been widely used in previous semantic analysis tasks \citep{pilehvar2019wic, bevilacqua2021recent} as our source of LKB. Then $D^t$, $\mathbf{c}$, and $t$ are jointly inputted to the text encoder with the following template. 
\begin{center}
    \{\textit{context}\} : \{\textit{$i$th sense's definition}\} 
\end{center}

C2D is computed by the inner product of the hidden representations on the definitions $\mathbf{d^t_{1...|D^t|}}$ and the context ${\mathbf{h}^c}^\intercal$. D2I is then calculated by the inner product of the hidden representations of the input images $\mathbf{h}^v_{1...|V^t|}$ and ${\mathbf{d}^t_{1...|D^t|}}^\intercal$. Both C2D and D2I input to the softmax function transformed into probability distributions. Then, we choose the image with the highest probability as the prediction. As a result, for the example in Figure \ref{fig:our_model_structure}, our model can predict the correct answer of the given context `Angora city', whereas the original CLIP wrongly selects an image of `Angora cat' that produced the highest probability (as shown in Figure~\ref{fig:clip_structure}), even though the network topology and the pretrained parameters in our model are the same as the original CLIP model. 

\section{Handling OOV with the Context-Aware Definition Generation}
\begin{figure}
    \centering
    \caption{Examples of GPT-3 generated definitions when \textcolor{dukeblue}{context}, \textcolor{burntorange}{target word}, and \textcolor{olive}{part-of-speech} are `\textcolor{dukeblue}{angora city}', `\textcolor{burntorange}{angora}' and `\textcolor{olive}{noun}' (\textcolor{olive}{n}) respectively.}
    \label{fig:GPT_gen_def}
    \begin{subfigure}[b]{\linewidth}
        \centering
         \includegraphics[width=\textwidth]{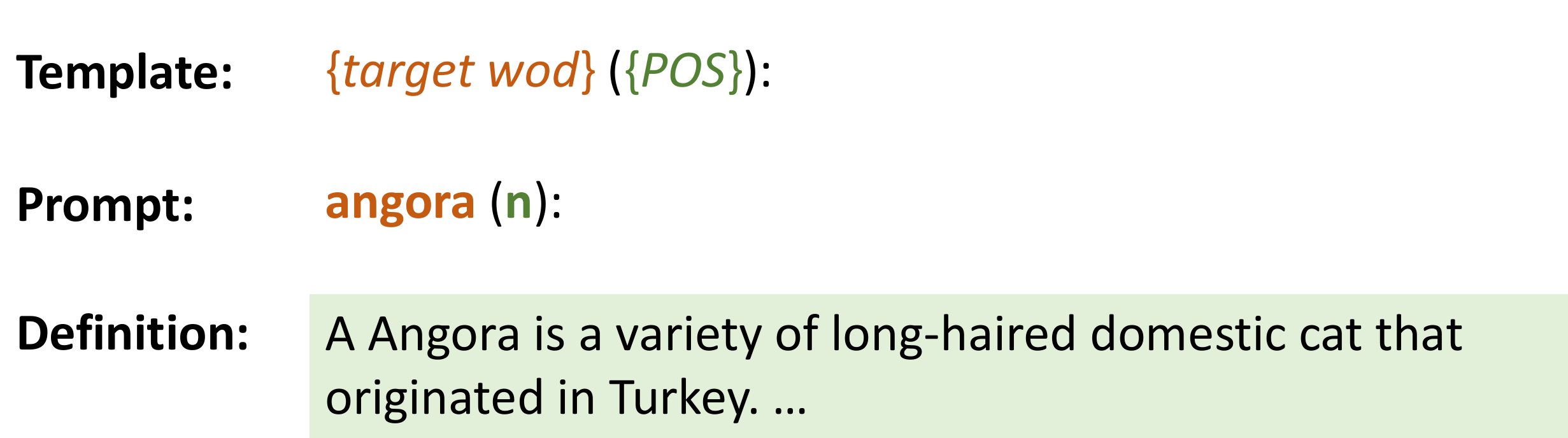}
         \subcaption{\citet{malkin2021gpt}'s definition generation.}
         \label{fig:def_gen_prev}
     \end{subfigure}
     \begin{subfigure}[b]{\linewidth}
        \centering
         \includegraphics[width=\textwidth]{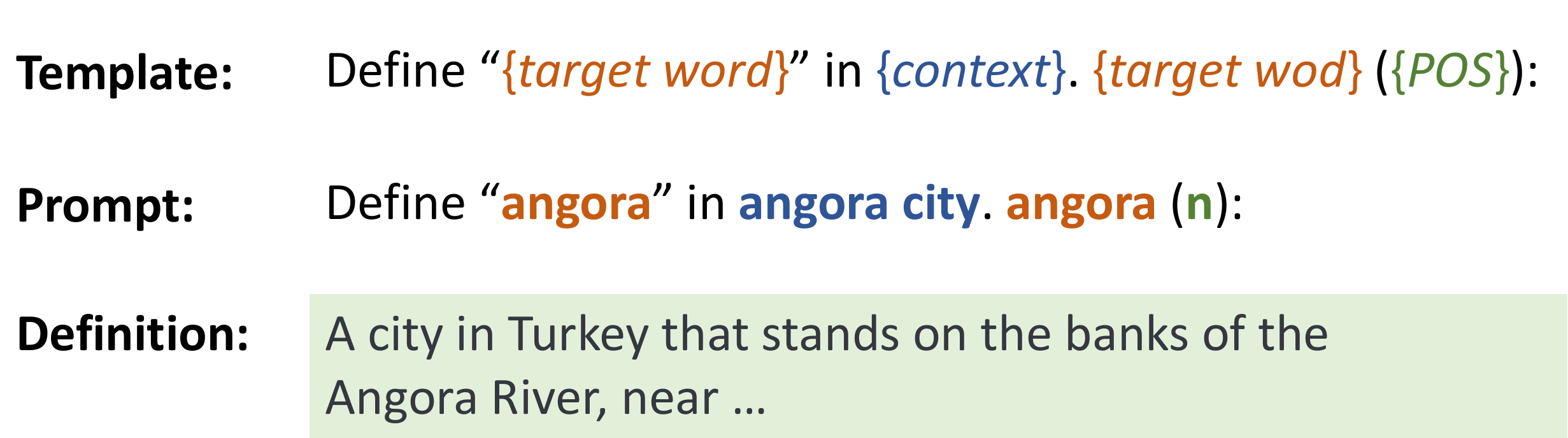}
         \subcaption{Our context-aware definition generation.}
         \label{fig:def_gen_ours}
     \end{subfigure}
\end{figure}

Not all words have their definitions available in a lexical knowledge-base. In particular, proper nouns, compound words, and foreign words frequently induce OOV issues. For example, in the SE23 dataset, about 14.33\% of target words' definitions are not found in the English WordNet. Therefore, we propose a solution to tackle the OOV issue with the definition generation approach. A previous study showed that GPT-3 can generate the definition of a novel word \citep{malkin2021gpt}. However, since this study does not consider the context of the word, it may not generate the definition in the correct sense. Thus, we suggest generating a definition with the prompt that considers both the context and the target word together.
%inspired by \citet{malkin2021gpt}'s approach that a definition of a new word can be generated by GPT-3. 

Figure~\ref{fig:GPT_gen_def} presents the generated definitions by the approach of \citet{malkin2021gpt} (Figure~\ref{fig:def_gen_prev}) and ours (Figure~\ref{fig:def_gen_ours}). Here, we add a conditional sentence that inputs the context of a target word. For example, when the target word is `angora' and the context is `angora city', we use a conditional sentence, ``Define \textcolor{burntorange}{``angora''} in \textcolor{dukeblue}{angora city.}'', in front of the previous input ``\textcolor{burntorange}{angora (\textcolor{olive}{n})}''. Indeed, in the example, the definition generated with our method shows a better description compared to the previous method. %Detailed examples of the definitions generated with the prompt are in Appendix~\ref{apx:Def_Gen}. 

% \begin{center}
%     \small
%     Define "\{\textit{target word}\}" in \{\textit{context}\}. \{\textit{target word}\} (\{\textit{POS}\}):
% \end{center}

\section{Experiments}
\subsection{Experimental Dataset}
\paragraph{SE23} We used the dataset in the SemEval-2023 Task 1 VWSD challenge \footnote{\url{https://semeval.github.io/SemEval2023/tasks.html}}\footnote{Note that, the data we use is the training set of the task. This is because the test set has a much smaller size of examples so it is advantageous to perform various case studies.}. It consists of 12,896 examples and 13,000 candidate images. Each example has 10 candidates that include 1 answer image and 9 distractors. Each context averagely contains 2.5 words. The dataset contains 14.33\% OOV words (1,845 out of 12,869). 

% \paragraph{VerSe} 

\subsection{Experimental Setting}
% Models, In the case of CLIP, among the three publicly available pretrained models (ViT-B/32, ViT-B/16, and ViT-L/14), we report the results of the model with the highest performance in SE23. Furthermore, we employ the FLAVA full model \footnote{\url{https://huggingface.co/facebook/flava-full}} in our experiments.
\paragraph{VWSD} For the experiments, we adopted two SOTA zero-shot ITM models, CLIP \footnote{\url{https://github.com/openai/CLIP}} and FLAVA \footnote{\url{https://huggingface.co/facebook/flava-full}}, as pretrained parameters are publicly available for both of them. Note that CLIP uses the text encoder and the image encoder at the same time while FLAVA contains the text encoder, the image encoder, and the multi-modal encoder. Herein, to calculate an image-text matching score, FLAVA uses the multi-modal encoder that cross-encodes image and text features simultaneously. In the case of calculating C2D, we exploit FLAVA's text encoder as the same as Figure~\ref{fig:our_model_structure}.

%On the other hand, 
We used WordNet 3.0 \footnote{https://www.nltk.org/howto/wordnet.html} as the main LKB. We also compare two GPT-3 generated definitions. The first one is \citet{malkin2021gpt}'s definition generation (DG). The other one is CADG (as described in Section 5). WN+CADG applies CADG's definitions in the case of OOV and uses WordNet definitions otherwise.

\paragraph{Definition Generation} 
% For all experiments with GPT-3 (generation and likelihood scoring), we used the davinci variant of the model, accessed through the free beta version of the API. In the generation of definitions from GPT-3, samples were taken with temperature 1 and truncated at the first line break or period. Because GPT-3 encountered dictionaries in its training data, acceptable samples were obtained for nearly all words.
%For all experiments, 
We re-implemented \citet{malkin2021gpt}'s definition generation experimental setting. Specifically, we sampled a definition for each example by utilizing GPT-3's Davinci variant which is known as the largest model and we generated samples with a temperature of 1.0.

\paragraph{Evaluation Criteria} Following \citet{raganato-etal-2023-semeval}'s setting, we evaluated VWSD models' performance with the hits at 1 (Hits@1) and the mean reciprocal rank (MRR). Moreover, we used Student's t-test \citep{student1908probable}, to verify the significance of differences in performance between models.   

\paragraph{Others} We prepared a pretrained WSD, T5$_{SemCor}$ \citep{wahle2021incorporating}. This model is a generative WSD model that a T5-large model \citep{2020t5} fine-tuned with SemCor \citep{raganato2017word}. Note that, SemCor is a large size word sense dataset annotated with the WordNet sense repository. Herein, we utilized the official checkpoint \footnote{\url{https://huggingface.co/jpwahle/t5-large-word-sense-disambiguation}}. In addition, we employed NLTK \citep{bird2009natural} to conduct word tokenization and part-of-speech tagging. All experiments were conducted on an NVIDIA A100 GPU with Ubuntu 22.04 version. 

\subsection{Experimental Results}

\begin{table}[]
\small
\begin{tabular}{@{ }c@{ }|l@{ }|cc}
\hline
 \multirow{2}{*}{Model} & \multirow{2}{*}{\begin{tabular}[c]{@{}l@{}}Source of\\ Definitions\\\end{tabular}} & \multicolumn{2}{c}{SE23} \\
\cline{3-4}
            &    & Hits@1                    & MRR\\
\hline\hline
\multirow{5}{*}{CLIP} & - & 73.00    &   82.72  \\
 & WN & 81.98 &   88.83\\
 & DG & 81.64 & 88.33 \\
 & CADG & 82.65 & 89.28 \\
 & WN+CADG & \textbf{83.08} & \textbf{89.60} \\
\hline
% \multirow{5}{*}{CLIP} & - & 73.00    &   82.72  &        &     \\
%  & WN & 81.98 &   88.83  &        &     \\
%  & DG & 81.64 & 88.33 &&\\
%  & CADG & 82.65 & 89.28 && \\
%  & WN+CADG & \textbf{83.08} & \textbf{89.60} & - & - \\
% \hline
\multirow{5}{*}{FLAVA} & - & 70.13 & 80.67    \\
 & WN & 78.34	&86.60 \\
 & DG & 74.05  &  84.49   \\
 & CADG & 75.13 &   84.53  \\
 & WN+CADG & \textbf{78.85} & \textbf{87.02}  \\
\hline
\end{tabular}
\caption{Experimental comparison on the ITM models with and without the gloss integration. There are three types of the source of definitions: 1) WordNet (WN) 2) \citet{malkin2021gpt}'s definition generation (DG) approach and 3) our context-aware definition generation (CADG). }
\label{tab:main_experiments}
\end{table}

The experimental results in Table~\ref{tab:main_experiments} show that the performances of CLIP and FLAVA are 73.00 and 70.13 on Hits@1, respectively. Incorporating definition descriptions of external LKB (WN) or generated (DG and CADG) significantly enhanced the performance in every experimental model. First, incorporating WordNet with our Bayesian style inference (WN) outperformed both of ITM models, 8.98\%p in CLIP ($p<1e-10$) and 8.72\%p ($p<1e-10$). DG and CADG also significantly improved performance in all cases ($p<1e-7$), but the increment in FLAVA was relatively lower than that of the CLIP. WN+CADG achieved the highest performance in both of CLIP and FLAVA.%The definitions generated by both DG and CADG approaches also significantly improve the performance of the original models in both models ($p<1e-7$). Meanwhile, the performance of our CADG achieved significantly higher performance compared to DG ($p<1e-3$). Finally, CADG+WN has marginally higher performance compared to CADG ($p>0.2$).

\begin{figure}
    \centering
    \begin{subfigure}[b]{\linewidth}
        \centering
         \includegraphics[width=\textwidth]{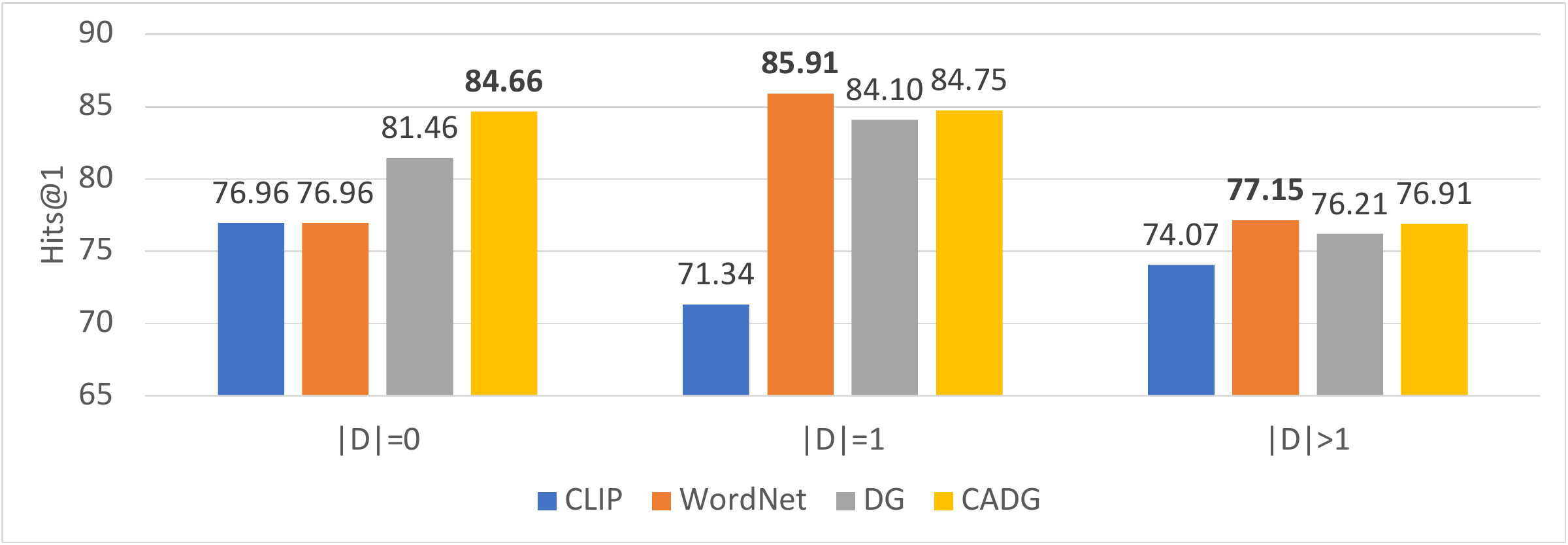}
         \subcaption{CLIP}
         \label{fig:clip_for_each}
     \end{subfigure}
     \begin{subfigure}[b]{\linewidth}
        \centering
             \includegraphics[width=\textwidth]{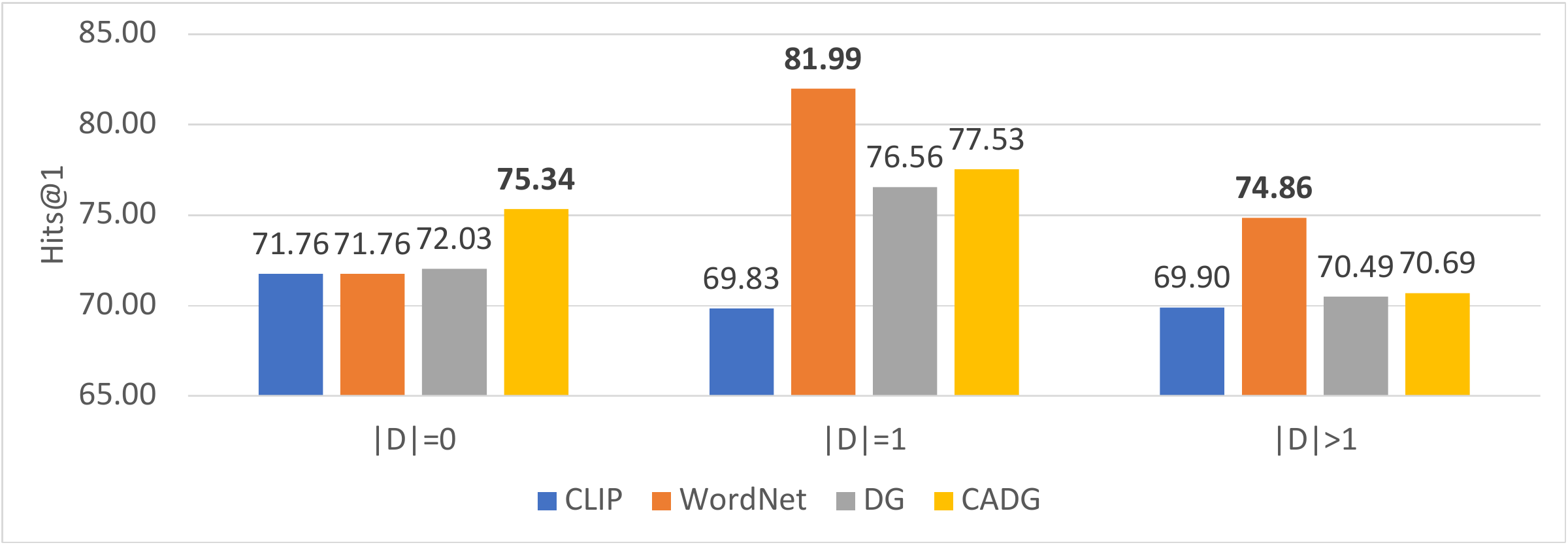}
         \subcaption{FLAVA}
         \label{fig:flava_for_each}
     \end{subfigure}
    \caption{Hits@1 score for the different number of definitions ($|D|$) of a target word in WordNet. }
    \label{fig:experiment_num_of_definition}
\end{figure}

On the other hand, to scrutinize the reasons for the performance improvements in more detail, we categorized examples into three categories according to the number of WordNet senses ($|D|$) of the target word. $|D|=0$ examples are target words with no entry in WordNet (OOV). $|D|=1$ examples are target words with only one sense in the WordNet (trivial). $|D|>1$ examples are target words with more than one sense in the WordNet (ambiguous).

 Figure~\ref{fig:experiment_num_of_definition} presents that incorporating WordNet definition enhanced the performance on ambiguous and trivial words in both of CLIP and FLAVA. In particular, the performance gain was remarkable in trivial words (from 71.34 to 85.91 and from 69.83 to 81.99 for CLIP and FLAVA, respectively). Moreover, even for ambiguous words, the performance is significantly improved ($p<1e-3$) without any additional training or the assistance of external systems such as WSD models. CADG substantially increased performance in both of OOV and trival words. Especially, when compared to DG, the performance differences are remarkable in OOV.

Meanwhile, while FLAVA shows prominent improvement via WordNet integration, the impact of generated definitions tends to be low compared to CLIP. Considering that WordNet definitions were manually constructed by experts, we speculate that this is because the model is sensitive to the quality of the input definitions.

% \begin{table}[]
% \small
% \begin{tabular}{l|cc|cc}
% \hline
%   & \multicolumn{2}{c}{S23} & \multicolumn{2}{c}{VerSe}  \\
% \cline{2-5}
%                 & Hits@1                    & MRR & Hits@1 & MRR  \\
% \hline\hline

% \hline

% \hline
% \end{tabular}
% \caption{Comparison }
% \end{table}

\section{Discussion}
\begin{table}[]
\small
\begin{tabular}{@{ }c@{ }|@{ }c@{ }|c@{ }|@{ }c@{ }}
\hline
$|D|$                      &  \# of Corrected & \# of Incorrected & Corrected Ratio \\
\hline\hline
2                         & 199           & 66            & 3.02  \\
3                         & 99            & 40            & 2.48  \\
4                         & 48            & 19            & 2.53  \\
5                         & 42            & 13            & 3.23  \\
6                         & 28            & 9             & 3.11  \\
7                         & 25            & 5             & 5.00  \\
8                         & 13            & 5             & 2.60  \\
9                         & 10            & 4             & 2.50  \\
10                        & 7             & 2             & 3.50  \\
10 \textless $|D|$        & 52            & 27            & 1.93  \\
\hline
total                     & 523           & 190           & 2.75  \\
\hline
\end{tabular}
\caption{Improvement rate on the ambiguous target words. $|D|$ is the ambiguity level of target words. `\# of Corrected' indicates the number of examples incorrect in CLIP but become correct in CLIP+WN. On the other hand, `\# of Incorrected' means the number of examples correct in CLIP but become incorrect in CLIP+WN. The corrected ratio is $\frac{\#~\text{of Corrected}}{\#~\text{of Incorrected}}$. }
\label{tab:ambiguous_target_analysis}
\end{table}

\subsection{Analysis on Ambiguous Target Words}
\label{sec:exp_C2D}
%\paragraph{Oracle Test}
%\paragraph{Incorporating Existing WSD Models}
We analyzed the performance change according to the ambiguity level of the ambiguous target word. Table~\ref{tab:ambiguous_target_analysis} presents the predictive change of the CLIP after incorporating WordNet. Herein, 523 examples go correct while 190 examples go incorrect. In particular, even in the case of highly ambiguous examples with $|D|$ greater than 10, the improvement rate is 1.93, and incorporating WordNet positively affects the performance improvement. These results are in line with previous research findings that ambiguous words can be recognized pre-trained LMs according to the given context \citep{gari2021let, kwon-etal-2022-medjex}. However, compared to the lower ambiguous cases, the performance improvement rate is lower. These results implies that enhancement for the highly ambiguous words are required.

\begin{table}[]
\small
\begin{tabular}{l|cc}
\hline
                & Hits@1                    & MRR \\
\hline\hline
-  & 74.07 & 82.72 \\
CLIP+WN & \textbf{77.15}   &  \textbf{88.83}  \\
\hline
%Not Relevant & & & 70.46& 80.35 \\
T5$_{SemCor}$ & 77.12 & 85.21    \\
% -  & 74.07 & 82.72 & 69.90 & 80.10  \\
% WN & 77.15   &  88.83 & 74.86 & 84.09 \\
% \hline
% %Not Relevant & & & 70.46& 80.35 \\
% T5$_{SemCor}$ & 77.12 & 85.21    \\
%BEM \cite{}         &                           &     &        &     \\
% \hline
% Oracle      & 81.87 &     &        &     \\
\hline
\end{tabular}
\caption{Experimental comparison of VWSD for the ambiguous target.}
\label{tab:comparison_with_pipeline}
\end{table}

Although WordNet integration improves performance for ambiguous target words, we still want to find out how competitive the performance improvement is. For this reason, we compared the performance of our WordNet-incorporated model with that of the pipeline system using the WSD model. To be specific, T5$_{SemCor}$, a finetuned WSD model, predicts WordNet sense in a given target word and context. The probability distribution for the candidate images was calculated based on the predicted sense.

Table~\ref{tab:comparison_with_pipeline} is the prediction result for ambiguous target words. Our model showed comparable results in the pipeline system and Hits@1 and achieved higher performance in MRR. This is due to the error cascading issue of pipeline systems \citep{finkel2006solving, kwon2019effective}. 
That is, in the pipeline system, errors in the WSD model directly lead to performance decrement. Otherwise, our approach is rather free from error cascading, since the C2D probability and the D2I probability work complementary to each other.

\subsection{Analysis on the Generated Definitions}
\begin{table}[]
    \small
    \centering
    \begin{tabular}{c|c|c}
    \hline
        Generation Method & Agreement & \# Examples\\
    \hline\hline
       DG  & 81.76& 159\\
       CADG  &\textbf{89.16} & 166\\
    \hline
    \end{tabular}
    \caption{Results of the human analysis on generated definitions.}
    \label{tab:def_gen_analysis}
\end{table}

% \begin{table}[]
%     \centering
%     \begin{tabular}{c|cc|cc}
%     \hline
%         &\multicolumn{2}{c|}{SE23}  & \multicolumn{2}{c}{VerSe}\\
%     \hline
%         Method & Hits@1 & MRR & Hits@1 & MRR  \\
%     \hline\hline
%         - & & & & \\
%        GD  & 75.50 & & & \\
%        CADG  & 80.50 & & & \\
%     \hline
%     \end{tabular}
%     \caption{Results of the human analysis on generated definitions.}
%     \label{tab:def_gen_analysis}
% \end{table}

\subsubsection{Evaluation on the Generated Definitions} 
In order to evaluate the quality of the generated definitions, we randomly sampled 200 examples from SE23 dataset. For each example, two annotators evaluated the (binary) agreement on the generated definitions with \citet{malkin2021gpt}'s approach (DG) and our approach (CADG). Inter-annotator agreement \cite{kvaalseth1989note} was $\kappa=0.625$. Finally, we only accept 159 examples of DG and 166 examples of CADG unanimously agreed by the annotators.

\begin{table*}[]
\center
\small
\begin{tabular}{@{} m{1cm}|l|l|p{4cm}|p{5.5cm}@{}}
\hline
                      & Target Word & Context & WordNet Answer Definition & Generated Definition \\
\hline\hline
\multirow{6}{*}{DG}   & give        & give communicate & convey or reveal information & to present something as a gift; to make a gift of something \\
\cline{2-5}
                      & landscape   & landscape genre  & painting depicting an expanse of natural scenery & A large area of land that can be seen from one place\\
\cline{2-5}
                      & fauve & fauve painter & a member of a group of French painters who followed fauvism & A fauve is a wild or undomesticated animal.\\
\hline
\multirow{5}{*}{CADG} & give & give communicate & convey or reveal information & to convey (information, etc.) \\
\cline{2-5}
                      & landscape & landscape genre & painting depicting an expanse of natural scenery & a genre of art that depicts natural scenery such as mountains, forests, rivers, and so on\\
\cline{2-5}
                      & fauve & fauve painter & a member of a group of French painters who followed fauvism & a French term meaning "wild beast," used to describe a group of early 20th-century ...\\
\hline
\end{tabular}
\caption{Examples of generated definitions with DG and CADG.}
\label{tab:definition_generation_examples}
\end{table*}

Table~\ref{tab:def_gen_analysis} represents the average human agreement scores on DG and CADG. The results show that our CADG achieved a higher performance compared to DG. Especially, in Figure~\ref{fig:GPT_gen_def} and Table~\ref{tab:definition_generation_examples}, we can find that the definitions of ambiguous words generated with CADG are semantically similar to that of the WordNet answer sense compared to DG, in line with the purpose for which it was designed.

\begin{table}[]
\small
\begin{tabular}{l|l|l|cc}
\hline
    Model   &  Agreement    & n  & Hits@1 & MRR \\
\hline\hline
CLIP      & -  & 159 &   71.70   &  82.29   \\
CLIP+DG   & Correct &  130   &    \textbf{83.85 }   &  \textbf{89.76}   \\
CLIP+DG   & Incorrect & 29 &   68.97     &    78.83 \\
\hline
CLIP      & -  & 166 &   68.67     & 79.78    \\
CLIP+CADG & Correct & 148   &  \textbf{82.43} &  \textbf{89.25}   \\
CLIP+CADG & Incorrect & 18 &    66.67   &  77.45   \\
\hline
\end{tabular}
\caption{VWSD performance according to the quality of generated definitions.}
    \label{tab:def_gen_VWSD_performance}
\end{table}

\subsubsection{Impact of the Generated Definitions' Quality} 
We also verified whether the quality of the generated definitions would affect the VWSD performance. Table~\ref{tab:def_gen_VWSD_performance} presents the experimental results on VWSD examples when we utilize the generated definitions that agreed (Correct) and disagreed (Incorrect) by the both annotators. Table~\ref{tab:def_gen_VWSD_performance} demonstrates that the quality of the generated definitions affects the performance of the downstream VWSD task indeed.
\begin{table}[]
\small
\begin{tabular}{c|ccc}
\hline
\multicolumn{1}{c|}{\multirow{2}{*}{}} & \multicolumn{3}{c}{n} \\
\cline{2-4}
\multicolumn{1}{c|}{} & 1 & 2 & 3 \\
\hline\hline
DG & 81.64 & \textbf{81.75} & 80.19 \\
\hline
CADG & 82.65 & \textbf{82.68} & 82.29 \\
\hline
\end{tabular}
\caption{The performance (Hits@1) according to the different number of sampled definitions (n).}
\label{tab:num_sampled_definitions}
\end{table}

\subsubsection{Experiments on Multiple Generated Definitions} 
Since we sampled a definition for each input example in main experiments, it is still questionable whether the number of sampled definitions affects the performance of the model. Table~\ref{tab:num_sampled_definitions} indicates the performance of DG and CADG according to the number of generated definitions (n) for each input. The results show that the number of sampled definitions is not significantly affecting the model's performance. To be specific, when the number of generated definitions is 2 for each input, the performance of DG and CADG increased by 0.09\%p and 0.03\%p respectively. 
Furthermore, when the number of generated definitions is 3, we can see that the performance even slightly decreases both DG and CADG. As a result, sampling multiple definitions for each input does not significantly affect performance or rather decreases performance.

\subsection{Error Analysis}
% Error analsis on WSD and VSD parts
% \begin{figure*}
%     \centering
%     \includegraphics[width=\textwidth]{Figures/VWSD_errors.pdf}
%     \caption{Illustrative concepts and an example input on a CLIP model \citep{radford2021learning}.}
%     \label{fig:VWSD_error}
% \end{figure*}

\subsubsection{VWSD}
Our model still suffers from error cascading from C2D probability though it is mitigated by the Bayesian style inference. The most typical error case is due to the error cascading in C2D probability calculation. Especially, due to the nature of neural networks \citep{guo2017calibration}, the over-confidence in the error classes frequently causes errors. For example, in Table~\ref{tab:vwsd_error}, we found that among the 10 senses of the target word `paddle' extracted from WordNet, the conditional probability for the correct sense was calculated as 0.00\%, resulting in an error in the final posterior calculation. Another error case is when there is no correct sense in WordNet. In the example, the target word `Thompson' indicates a firearm, but WordNet contains only personal information. This is a separate issue from OOV with no entry for the target word, and we observed that it mainly occurs in proper nouns.

\begin{table}[]
\footnotesize
\begin{tabular}{@{}c@{ }|@{ }c@{ }|@{ }l@{ }|@{ }c@{}}
\hline
Target Word & Context & \multicolumn{1}{c|@{ }}{WordNet Definitions} & Probs. \\
\hline\hline
\multirow{2}{*}{paddle} & \multirow{2}{*}{paddle beat} & walk unsteadily & 99.95\%\\
\cline{3-4}
                        & & give a spanking to & 0.00\% \\
\hline
\multirow{2}{*}{Thompson} & \multirow{2}{*}{\begin{tabular}[c]{@{}l@{}}Thompson \\ 
submachine\end{tabular}} & \begin{tabular}[c]{@{}l@{}}United States classical \\ archaeologist…\end{tabular}  & 0.00\% \\
\cline{3-4}
& & \begin{tabular}[c]{@{}l@{}}English physicist \\ (born in America) …\end{tabular}   & 100.00\% \\
\hline
\end{tabular}
\caption{Examples of VWSD error cases. Probs. stands for $P(D^t_i|\mathbf{c},t)$.} 
\label{tab:vwsd_error}
\end{table}

\subsubsection{Definition Generation}

\begin{figure}
    \centering
    \begin{subfigure}[b]{\linewidth}
        \centering
         \includegraphics[width=\textwidth]{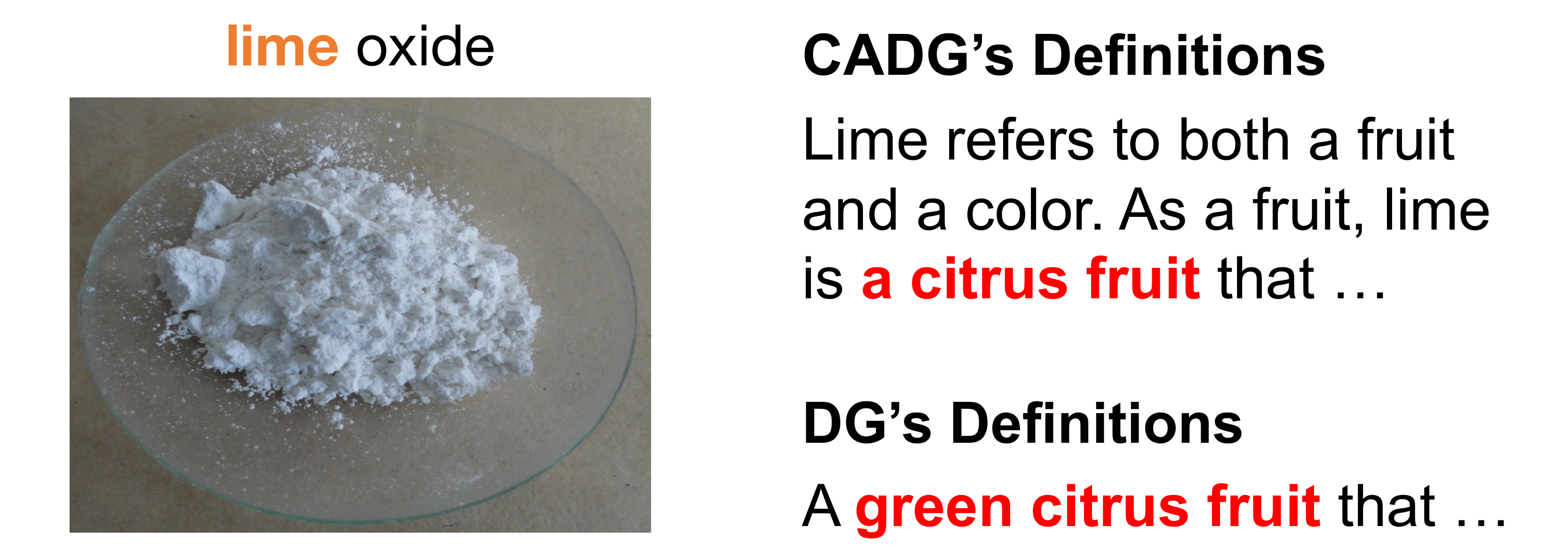}
         \subcaption{An example of the misdisambiguation}
         \label{fig:misdisambiguation}
     \end{subfigure}
     \begin{subfigure}[b]{\linewidth}
        \centering
         \includegraphics[width=\textwidth]{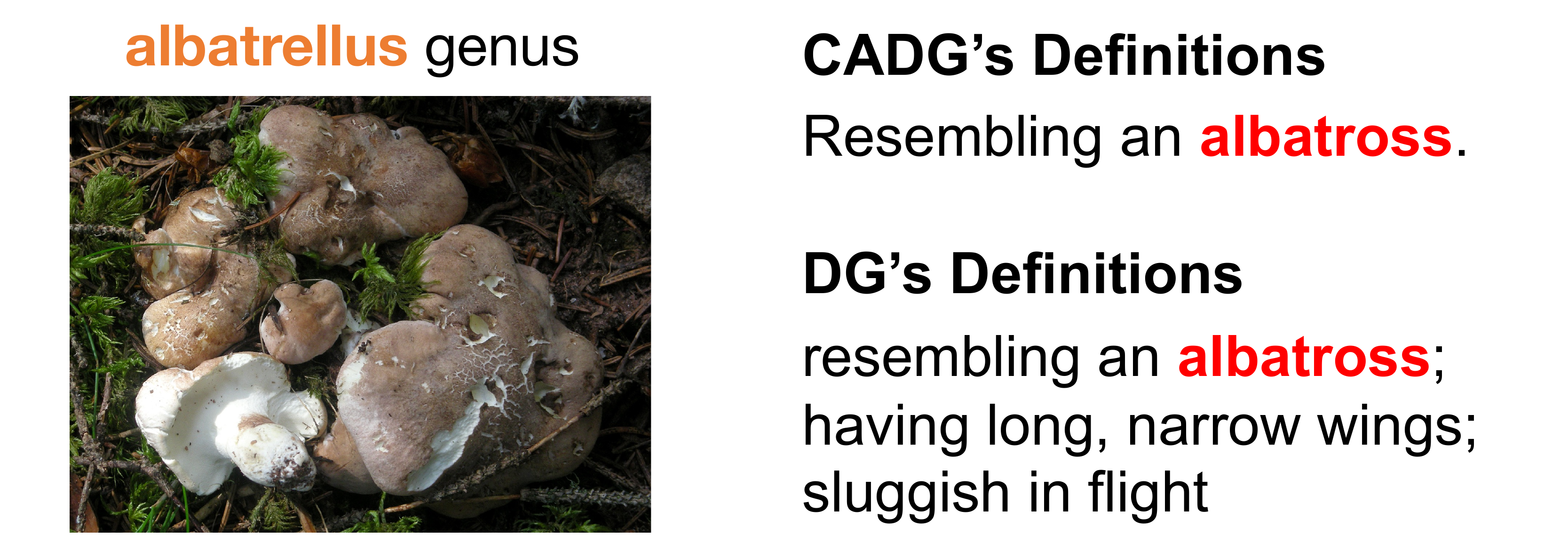}
         \subcaption{An example of the hallucination}
         \label{fig:hallucination}
     \end{subfigure}
    \caption{Examples of incorrectly generated definitions. }
    \label{fig:def_gen_error}
\end{figure}

We found two representative error cases in the results of the definition generations: 1) misdisambiguation and 2) hallucination. The misdisambiguation is when the GPT3 generates the polysemy's definition. In Figure~\ref{fig:misdisambiguation}, considering the context of ``lime oxide'', we would expect a definition of \textcolor{burntorange}{lime stone} to be generated. However, we can notice that both approaches generate a definition for \textcolor{red}{lime fruit}. On the other hand, as pointed out in previous research \citep{ishii2022survey}, we also observed that GPT3 generates hallucinations. Figure~\ref{fig:hallucination} is an example of the hallucination issue. \textcolor{burntorange}{albatrellus} which is a type of a fungi in the context of ``albatrellus genus,'' nevertheless the definitions generated by both approaches are pertaining to the albatross, a species of bird. Detailed examples of error cases can be found in Appendix~\ref{apx:err_definitions}.

\section{Conclusion and Future Work}
This paper introduces a novel VWSD methodology to effectively incorporate gloss information from an external resource. Our work mainly has two innovations: 1) Bayesian style inference for SOTA ITMs, and 2) Context-aware definition generation with GPT-3 to overcome the OOV issue. Experimental results show that our proposed Bayesian style inference-based WordNet integration significantly improves VWSD performance without additional training. For the ambiguous target words, the performance of our approach is comparable to pipeline systems using finetuned WSD models. Moreover, context-aware definition generation helps mitigate OOV issues in the downstream VWSD tasks and shows higher performance compared to the previous definition generation approach.

In the future, we plan to tackle the error cascading caused by over-confidence in C2D probability. For this, we may explore a prompting that is known to have good performance in zero-shot prediction \citep{liu2023pre}. In addition, to deal with the hallucination and misdisambiguation problems of GPT-3 generated definitions, we may employ controllable generation by resampling \citep{ji2022survey}.

\label{sec:definition_generation_err}

\section*{Limitations}
Our work has the following limitations. First, we only used one evaluation data, namely SE23, because it is the only data suitable for the VWSD setting, especially for the OOV examples. In addition, our methodology relies entirely on WordNet. Therefore, this may be limited the model's ability when the target word is a proper noun such as a named entity. Finally, we depend on the results of GPT-3 definition generation to handle OOV words. Since the generated definitions may contain errors, as revealed in the qualitative analyses, the errors led to incorrect predictions.
%performance evaluation data. This is because data
 %Instead, we scrutinized the pros and cons of the proposed methods through exhaustive qualitative analyses. 
%Nevertheless, it can be necessary to add other data necessary to verify our hypotheses in the future.

%However, as revealed in the qualitative analysis, 

\section*{Ethical Consideration}
The generated definitions were annotated by two annotators. Both annotators were fully paid by complying with local minimum wage regulation.
% we paid them by obeying the minimum wage standard of national law
In addition, in the sampled definition generations, the authors could not find any statements violating the ACL anti-harassment policy. However, generated definitions that authors have not vetted are still at risk of containing toxic or hates contents (e.g. racism, insulting or xenophobic). %Finally, we are sharing the experimental source code with the reviewers. The analysis results and the source code will be available to the public after anonymity period.

\section*{Acknowledgement}
Research reported in this study was in part supported by the Center of Biomedical and Health Research in Data Sciences (CHORDS) in UMass Lowell. 

\bibliography{custom}
\bibliographystyle{acl_natbib}

\newpage
\onecolumn
\appendix

\section{Case Study on Incorrectly Generated Definitions}
\label{apx:err_definitions}

Table~\ref{tab:DG_errs} and Table~\ref{tab:CADG_errs} present the all incorrectly generated definitions that described in Section~\ref{sec:definition_generation_err}. Herein, we found the following three error types: 1) Misdisambiguation, 2) Hallucination, and 3) Others. 

First of all, the misdisambiguation cases are caused by bias in the pretraining, and we can notice that CADG has less misdisambiguation compared to DG. Especially, we can see that GPT-3 generated more than one definitions of the target words `conch', `reaper', and `ruin' in DG, while we could not found such cases in our approach.  On the other hand, hallucination cases are when the generated definitions are definitions of completely different terms with similar spellings (`stonechat' of CADG, `driftfish' of DG), or cases in which the detailed descriptions are incorrect although they are somewhat similar (`osteostraci' of CADG, `nestor' of DG). Especially, in Table~\ref{tab:DG_errs} of `wulfenite' and `cordierite,' we can notice that definitions are generated with parts of each lexicon ("wulfen," and "cord"). Finally, in other cases, the generated definitions may not be in definition form (`lynching' of CADG, `areca' of DG), or the contents of the target word is output as itself (`wulfenite' of CADG).

% \usepackage[normalem]{ulem}
\begin{table}[!ht]
\scriptsize
\begin{tabular}{@{ }c|c|p{4.5cm}|c|c|c|c@{ }}
\hline
context & target word & GPT Definition & Agreement & Misdisambiguation & Hallucination & Others \\
\hline\hline
niche condition & niche & A niche is a situation or activity that is specially suited to a person's interests, abilities, or character. & FALSE & \checkmark & & \\
lighterage transfer & lighterage & the cost of using a lighter (a small ship used to transfer cargo from a larger ship to shore) to transfer cargo & FALSE & \checkmark & & \\
lynching practice & lynching & What is an example of a hate crime? & FALSE & & & \checkmark \\
stonechat chat & stonechat & a chatty person who talks excessively about trivial things & FALSE & & \checkmark & \\
wheatear thrush & wheatear & of a pale yellowish-brown color, like that of wheat & FALSE & & \checkmark & \\
thank convey & thank & Thank you for your help. & FALSE & & \checkmark & \\
driftfish butterfish & driftfish & moving aimlessly or erratically; drifting. & FALSE & & \checkmark & \\
fumble blooper & fumble & fumble (verb): to make a mistake or do something awkwardly & FALSE & \checkmark & & \\
snoek table-fish & snoek & clad in snow; frosty. & FALSE & & \checkmark & \\
variation dance & variation & a repeated section of music in which the melodic, rhythmic, and/or harmonic content is altered each time it occurs & FALSE & \checkmark & & \\
stationery paper & stationery & of or relating to the writing and supplies used for writing by hand & FALSE & \checkmark & & \\
lime oxide & lime & lime refers to both a fruit and a color. As a fruit, lime is a citrus fruit that is used to add a sour or acidic flavor to food. & FALSE & \checkmark & & \\
behemoth anomaly & behemoth & a massive and powerful thing, often one that is difficult to control or deal with & FALSE & \checkmark & & \\
sic carbide & sic & intentionally so written (used after a printed word or phrase to indicate that the spellings, punctuation, capitalization, etc., have been retained from the original). & FALSE & \checkmark & & \\
wulfenite mineral & wulfenite & containing or consisting of wulfenite mineral. & FALSE & & & \checkmark \\
osteostraci order & osteostraci & a subclass of fish that includes the lampreys and hagfishes & FALSE & & \checkmark & \\
ramontchi shrub & ramontchi & a small, evergreen shrub of the genus Rhamnus, native to Japan and Korea. The ramontchi shrub is grown for its ornamental value, and its small, dark green leaves are used in traditional medicine. & FALSE & & \checkmark & \\
nestor genus & nestor & A person who is very old or who has lived for a long time & FALSE & & \checkmark & \\
\hline
\end{tabular}
\caption{Error cases of our CADG}
\label{tab:CADG_errs}
\end{table}

\begin{table}[!h]
\scriptsize
\begin{tabular}{@{ }c|c|p{4.5cm}|c|c|c|c@{ }}
\hline
Context & Target word & GPT Definition & Agreement & Misdisambiguation & Hallucination & Others \\
\hline\hline
niche condition & niche & A distinct segment of a market, usually populated by a group of people with specific needs or interests. & FALSE & \checkmark & & \\
lame fabric & lame & a person or animal that is lame in one or more legs & FALSE & \checkmark & & \\
razz mock & razz & a type of teased or tangled hair & FALSE & & & \\
burn turn & burn & A burn is a type of injury to the skin or other tissues caused by heat, electricity, chemicals, or radiation. & FALSE & \checkmark & & \\
bursera genus & bursera & A burser is a small sac or saclike cavity containing synovial fluid, located between a bone and a tendon or between a bone and the skin. & FALSE & & \checkmark & \\
mara rodent & mara & A small, gazelle-like antelope that is found in Africa. & FALSE & \checkmark & & \\
cordierite mineral & cordierite & Resembling a cord or rope; twisted; rope-like. & FALSE & & \checkmark & \\
distemper painting & distemper & a disease of domesticated animals, especially dogs, caused by a virus and characterized by fever and gastrointestinal and nervous system disorders & FALSE & \checkmark & & \\
pilate procurator & pilate & A type of exercise that uses slow, controlled movements to improve posture and flexibility. & FALSE & & \checkmark & \\
driftfish butterfish & driftfish & serene, drifting along peacefully & FALSE &  & \checkmark & \\
areca palm & areca & (adj.) & FALSE & & & \checkmark \\
spiceberry shrub & spiceberry & A spiceberry is a flowering plant in the genus Spirea, native to North America and Asia. & FALSE & & \checkmark & \\
conch gastropod & conch & 1. a spiral shell of certain marine Gastropoda 2. the shell of a Strombus, used as a trumpet by certain Native American peoples 3. in Buddhism, a spiral shell that is blown as a horn, used to summon beings to assembly & FALSE & \checkmark & &  \\
reaper agro-technology & reaper & 1. Having a disastrous effect; causing ruin or destruction. 2. (Of a person) Bringing death or misfortune. & FALSE & \checkmark & &  \\
packera genus & packera & A person who packs or wraps something, typically for shipping. & FALSE & & \checkmark & \\
variation dance & variation & the act, process, or result of varying in quality, quantity, or degree; diversity: the variations in color of a parrot's plumage. & FALSE & \checkmark & & \\
mamma exocrinopathy & mamma & A mamma is a mother. & FALSE & \checkmark & & \\
give communicate & give & to present something as a gift; to make a gift of something & FALSE & \checkmark & & \\
landscape genre & landscape & A large area of land that can be seen from one place & FALSE & \checkmark & & \\
passer genus & passer & a person who is authorized to pass or go through something, especially a military checkpoint & FALSE & \checkmark & & \\
stationery paper & stationery & of or relating to stationery or the office supplies used for writing and printing & FALSE & \checkmark & & \\
calypso orchid & calypso & A style of music originating in Trinidad and Tobago that is characterized by a heavy rhythm, often created with drums, guitars, and other percussion instruments. & FALSE & \checkmark & & \\
lime oxide & lime & A green citrus fruit that is used to add flavor to food and drinks. & FALSE & \checkmark & & \\
sic carbide & sic & Meaning "so" or "very," sic is derived from the Latin adverb sic, meaning "thus" or "just as." & FALSE & \checkmark & & \\
wulfenite mineral & wulfenite & relating to or resembling a wolf & FALSE & & \checkmark & \\
ramontchi shrub & ramontchi & Ramontchi is a type of fish found in the rivers of southern Japan. It is prized for its delicate flavor and is often used in sushi. & FALSE & & \checkmark & \\
nestor genus & nestor & a mentor or guide, especially one who is older or more experienced & FALSE & & \checkmark & \\
ruin destruction & ruin & 1. the remains of a building or city, typically one that is in ruins 2. a person or thing that is severely damaged or destroyed 3. a person's career, reputation, or life being ruined & FALSE & \checkmark & &  \\
pleiades nymph & pleiades & A group of seven stars in the constellation Taurus, typically visible to the naked eye. Also called the Seven Sisters. & FALSE & \checkmark & & \\
\hline
\end{tabular}
\caption{Error cases of DG \citep{malkin2021gpt}}
\label{tab:DG_errs}
\end{table}

% \section{Details on the Definition Generation}
% \label{apx:Def_Gen}
% For example, 

% \subsection{Examples of Generated Definitions}

\end{document}